\documentclass[letterpaper, 10 pt, conference]{ieeeconf}  

\IEEEoverridecommandlockouts                              

\usepackage[utf8]{inputenc}
\usepackage[T1]{fontenc}
\usepackage{url}
\usepackage{cite}
\usepackage{microtype}
\usepackage{amsmath}
\usepackage{amssymb}
\usepackage{amsfonts}
\usepackage{nicefrac}
\usepackage{graphicx}
\graphicspath{{images/}}
\usepackage[table]{xcolor}
\usepackage{booktabs}
\usepackage{tabularx}
\usepackage{multirow}
\usepackage{makecell}
\usepackage{array}
\usepackage{ragged2e}
\usepackage{algorithm}
\usepackage{algpseudocode}

\usepackage[hidelinks]{hyperref}

\newcommand{\up}[1]{\textcolor{green!55!black}{$\uparrow$#1}}
\newcommand{\down}[1]{\textcolor{red!70!black}{$\downarrow$#1}}
\newcommand{\same}{\textcolor{gray!70}{--}}

\title{\LARGE \bf
RoboNeuron: A Middle-Layer Infrastructure for Agent-Driven Orchestration in Embodied AI
}

\author{
    Weifan Guan$^{1,2}$, Qinghao Hu$^{1,\dagger}$, Huasen Xi$^{1,2}$, Chenxiao Zhang$^{1,2}$, Aosheng Li$^{2,3}$, and Jian Cheng$^{1,3,4,\dagger}$
    \thanks{$^{\dagger}$Corresponding authors.}
    \\[0.6em]
    $^{1}$Institute of Automation, Chinese Academy of Sciences \\
    $^{2}$University of Chinese Academy of Sciences \quad $^{3}$AiRiA \quad $^{4}$MAICRO
    \\[0.5em]
    {\tt\small \{guanweifan2024, huqinghao2014\}@ia.ac.cn, jcheng@nlpr.ia.ac.cn}
}

\begin{document}

\maketitle
\thispagestyle{empty}
\pagestyle{empty}

\begin{abstract}

Vision-language-action (VLA) models and LLM agents have advanced rapidly, yet reliable deployment on physical robots is often hindered by an interface mismatch between agent tool APIs and robot middleware. Current implementations typically rely on ad-hoc wrappers that are difficult to reuse, and changes to the VLA backend or serving stack often necessitate extensive re-integration. We introduce RoboNeuron, a middleware layer that connects the Model Context Protocol (MCP) for LLM agents with robot middleware such as ROS2. RoboNeuron bridges these ecosystems by deriving agent-callable tools directly from ROS schemas, providing a unified execution abstraction that supports both direct commands and modular composition, and localizing backend, runtime, and acceleration-preset changes within a stable inference boundary. We evaluate RoboNeuron in simulation and on hardware through multi-platform base control, arm motion, and VLA-based grasping tasks, demonstrating that it enables modular system orchestration under a unified interface while supporting backend transitions without system rewiring. The full code implementation of this work is available at github repo :\href{https://github.com/guanweifan/RoboNeuron}{RoboNeuron}.

\end{abstract}

\section{INTRODUCTION}

Vision-language-action (VLA) models and agent systems have progressed quickly in recent years, showing strong performance in language understanding, visual perception, and action generation. However, deploying these systems reliably on real robots remains difficult due to infrastructural bottlenecks: agent ecosystems operate via tool calls, whereas robots expose capabilities through middleware interfaces and streaming I/O. On the agent side, interactions are usually expressed as tool calling, where perception, inference, and control are triggered through structured interfaces. Conversely, robot capabilities are managed by middleware and messaging ecosystems. Because these two environments are not naturally aligned, accessing robot capabilities is often difficult to standardize across projects, and system iteration is frequently constrained by low-level engineering details.

In practice, this gap manifests as three primary challenges. First, scaling the interface bridge is difficult because robot capabilities are often exposed through heterogeneous interfaces. Converting these into agent-callable tools via manual wrappers increases maintenance costs and limits reuse across different projects. Second, the lack of a composable execution abstraction makes it hard for agents to organize tasks of varying complexity, from simple tool calls to multi-stage perception and control sequences. Finally, different VLA projects usually come with their own deployment code and serving interfaces. Moving from one model or inference stack to another often requires new wrappers and repeated integration work, which raises switching cost and makes comparisons harder to attribute to model differences alone.

\begin{figure*}[t]
\centering
\includegraphics[width=0.50\textwidth]{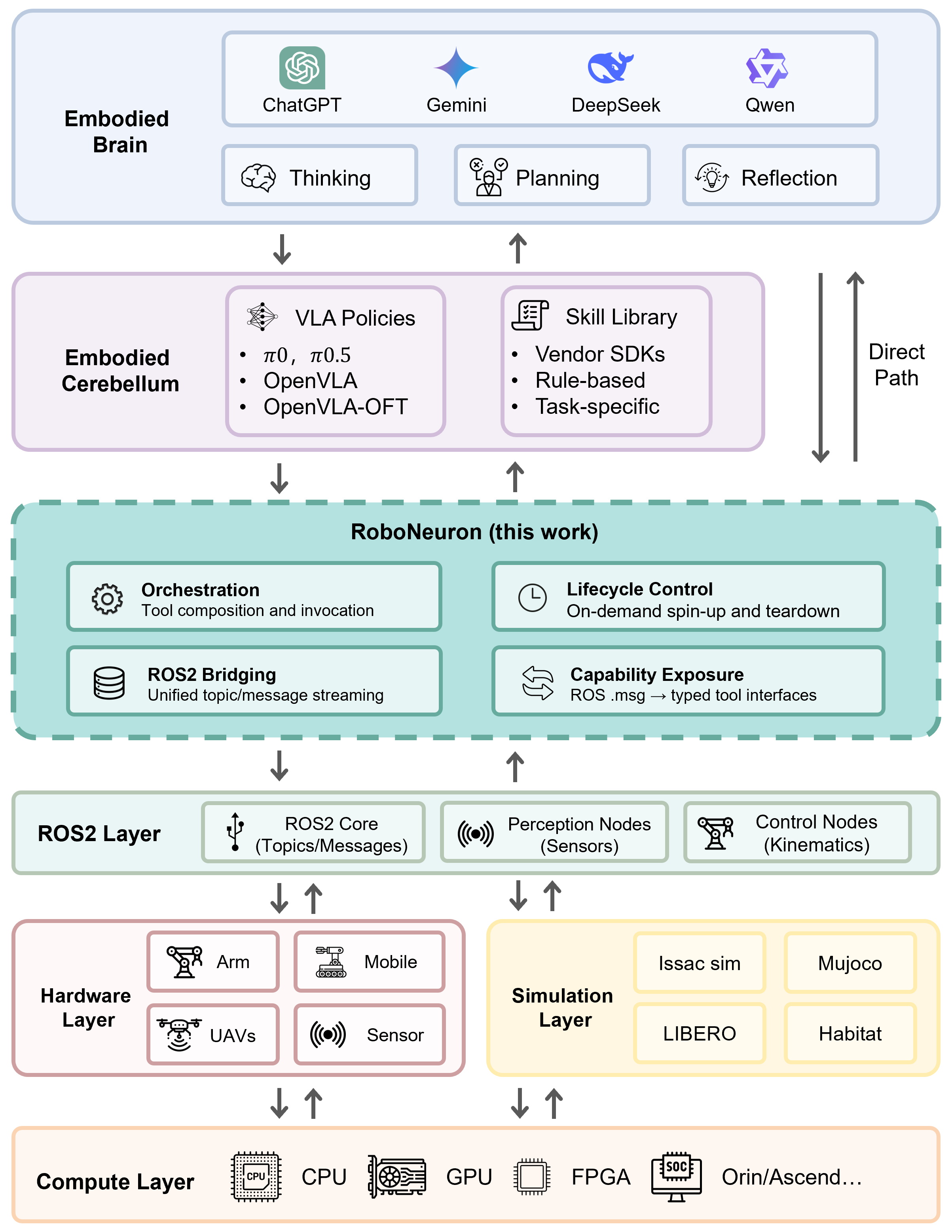}
\caption{Position of RoboNeuron in the broader embodied-agent stack. RoboNeuron sits between high-level agent reasoning and policy/skill layers above, and the ROS2, hardware, simulation, and compute layers below. In this position, it provides orchestration, lifecycle control, ROS2 bridging, and capability exposure, while preserving a direct path to lower-level execution resources when needed.}
\label{fig:stack}
\end{figure*}

To address these issues, we present RoboNeuron, a middleware layer between the agent ecosystem and the robotics ecosystem that connects MCP and robot middleware such as ROS2. RoboNeuron exposes robot-side interfaces as MCP tools by packaging ROS messages into structured, callable definitions. It employs a schema-based conversion mechanism to automatically derive tools from ROS message definitions, which reduces the manual effort required to update interfaces. Built on this unified interface, RoboNeuron organizes task execution into ROS message tools and perception, inference, and control modules. This allows an agent to choose between two execution modes: direct calls to low-level ROS tools for imperative control, or the construction of an execution path by composing specialized modules. VLA-related changes are contained within the inference module, creating a stable boundary that supports fast backend swapping. Consequently, RoboNeuron integrates multiple VLA backends and acceleration configurations without altering the overall module topology. This approach limits the impact of system changes to a well-defined boundary, reducing repeated integration work and lowering the cost of performance comparisons. Figure~\ref{fig:stack} places RoboNeuron in the broader embodied-agent stack and highlights its role between agent-side reasoning, policy/skill layers, and ROS2-based robot execution.

RoboNeuron is a middleware infrastructure that bridges agent tool calling and robot middleware, rather than a planning algorithm or a replacement for existing robot control stacks. Its contributions are threefold:

\begin{enumerate}
\item \textbf{Unified tool interface with scalable schema-based bridging.} We propose a derivation pipeline that generates agent-callable tool signatures from robot-side message and interface schemas, keeping agent-facing interfaces aligned with robot I/O as APIs evolve.
\item \textbf{Composable modular execution with agent-selectable paths.} We introduce an execution abstraction that lets agents coordinate tasks through a single interface by choosing between a direct path for low-latency primitives and a perception--inference--control (PIC) composition for persistent closed-loop behaviors.
\item \textbf{Topology-preserving backend and acceleration switching.} By localizing VLA-specific logic within a stable inference boundary, RoboNeuron supports switching VLA backends, inference runtimes, and acceleration presets without changing the surrounding communication topology.
\end{enumerate}

\section{RELATED WORK}
\label{sec:related}

\paragraph{\textbf{LLM Agents for Robotics}}
Prior research has used language models and LLM agents to organize task-level reasoning and execution on robots. Examples include skill-selection systems such as SayCan\cite{SayCan}, program-structured control such as Code as Policies\cite{CodeAsPolicies} and ProgPrompt\cite{ProgPrompt}, and integrated systems such as AutoRT\cite{AutoRT}, OK-Robot\cite{OKRobot}, and Inner Monologue\cite{InnerMonologue}. Other work, such as VoxPoser\cite{VoxPoser}, further studies how language can be grounded into robot actions and manipulation behavior. Together, these studies show that language models can help with skill choice, task decomposition, grounding, and multi-step embodied execution. However, their main focus is usually on planning logic, reasoning structure, or end-to-end agent behavior rather than reusable runtime infrastructure. In practice, robot I/O, tool exposure, and execution control are still often handled through project-specific code, which makes it difficult to reuse the same interface and execution semantics across systems. RoboNeuron instead targets the runtime layer beneath task-level reasoning, with the goal of making agent tool calling align more cleanly with robot-side execution under a unified capability interface.

\paragraph{\textbf{Vision-Language-Action Models}}
VLA models map visual observations and language instructions to robot actions. This includes multi-task policy learning frameworks such as RT-1\cite{RT1} and RT-2\cite{RT2}, open-source efforts such as OpenVLA\cite{OpenVLA}, OpenVLA-OFT\cite{OpenVLA-OFT} and the $\pi$ series\cite{pi0,pi05,pi06}. Related data efforts such as Open X-Embodiment\cite{OpenXEmbodiment} further highlight the push toward broader and more reusable embodied policy training. At the same time, VLA systems differ not only in model design but also in action representations, serving setups, inference organization, and deployment code, and these choices strongly affect deployment cost and efficiency\cite{EfficientVLA}. As a result, moving from one model or inference stack to another often requires new wrappers, new configuration logic, and repeated integration work around the model itself. RoboNeuron addresses the runtime-layer part of this problem by introducing a stable inference boundary that localizes VLA-related changes, so that backend and acceleration switching can be done without changing the surrounding system topology.

\paragraph{\textbf{Robot Middleware and Tooling}}
Robotic software benefits from the mature ROS2\cite{ROS2Architecture} ecosystem, where middleware abstractions support modular development. Widely used components such as MoveIt\cite{MoveIt} and Nav2\cite{Nav2}, together with behavior-tree-based organization\cite{BehaviorTreesSurvey}, provide reusable building blocks for perception, planning, navigation, and control. More broadly, middleware systems such as OpenRTM\cite{OpenRTM} also reflect a long-standing effort to make robot software modular and composable. These tools are effective developer-facing abstractions, but they are not designed as agent-facing callable interfaces. LLM agents instead require structured, discoverable, and stable tool schemas that can serve as a uniform entry point for invocation. This creates a gap between mature robot-side middleware abstractions and the interface form expected by agent tool calling. RoboNeuron builds directly on ROS2 to bridge that gap: it derives agent-callable tools from robot-side schemas and keeps those interfaces aligned with the underlying robot I/O and execution stack.

\paragraph{\textbf{Agent-Robot Bridging Infrastructures}}
Several systems focus on exposing robot interfaces as callable tools for agents. The MCP specification\cite{MCPspec} provides a common protocol for tool calling, and implementations such as ROS2mcpserver\cite{ros2mcpserver} and ROS2mcp\cite{ros2mcp} show that ROS capabilities can be connected to such interfaces in practice. These systems are useful for basic connectivity, but they mainly stop at protocol bridging and do not fully address scalable schema derivation, lifecycle-aware long-running execution, or reuse across different execution patterns. Other platform-style frameworks, such as RynnRCP\cite{rynnrcp} and RoboOS\cite{roboos}, explore broader architectures for robot skills, distributed execution, or cloud-connected robotics. These projects help motivate a more systematic software layer for embodied systems, but they target a wider system scope than the one considered here. RoboNeuron takes a narrower middle-layer position: it does not propose a new operating system or task-planning framework, but instead focuses on unified tool exposure, composable execution paths, and topology-preserving backend switching directly on top of ROS2.

\section{METHOD}
\label{sec:method}

\begin{figure*}[t]
\centering
\includegraphics[width=0.7\textwidth]{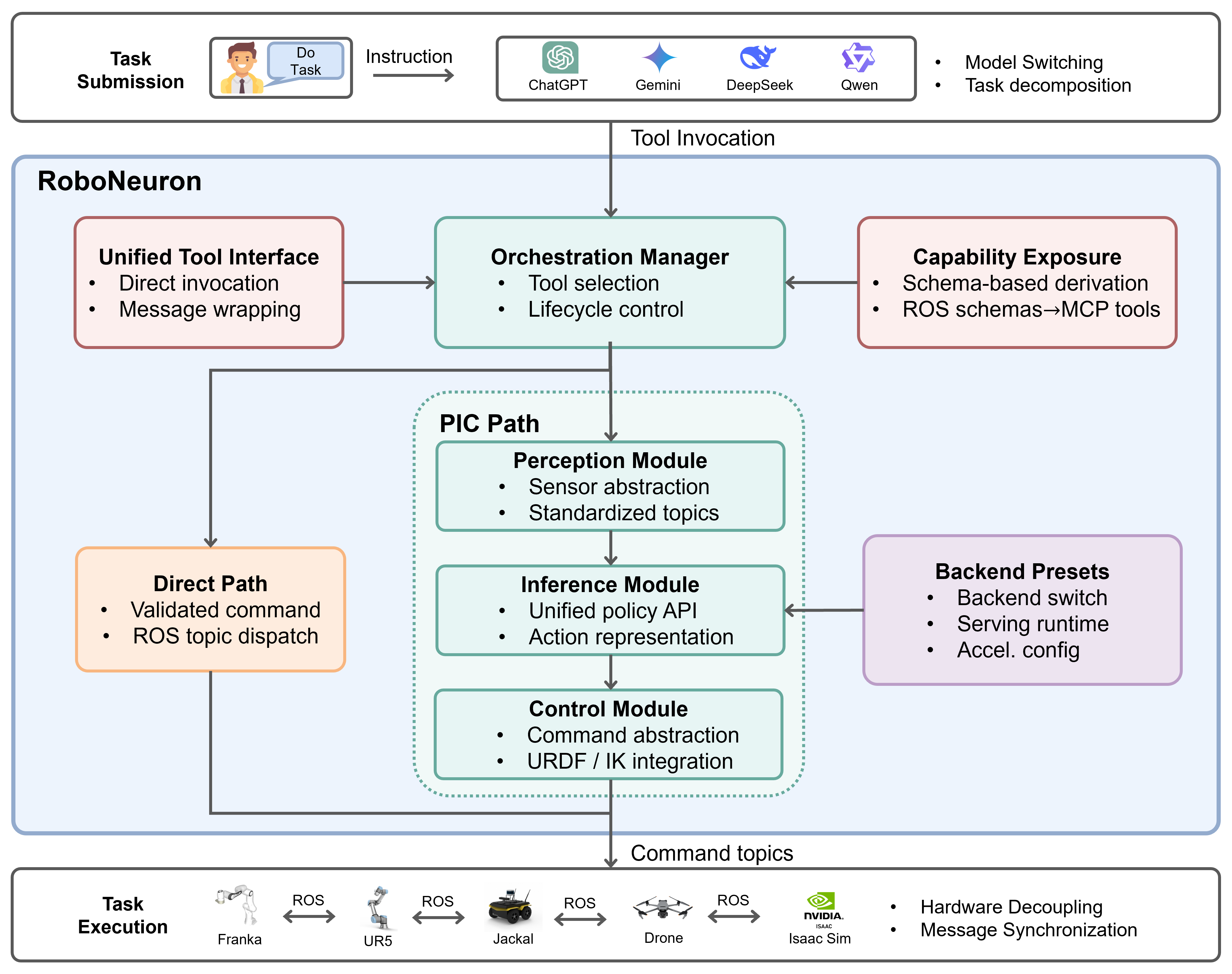}
\caption{Internal architecture of RoboNeuron. User instructions are turned into tool invocations that enter the orchestration manager through a unified tool interface, while robot-side capabilities are exposed as MCP tools through schema-based derivation from ROS schemas. RoboNeuron supports two execution paths: a direct path for validated command dispatch and a PIC path for perception--inference--control composition. Backend presets are injected only at the inference module, while command topics and downstream task execution remain unchanged.}
\label{fig:framework}
\end{figure*}

RoboNeuron serves as a middleware infrastructure for agentic VLA deployment, bridging agent-side tool calling with robot-side streaming execution. The architecture decouples semantics from transport by bifurcating the system into a control plane and a data plane. The control plane manages a unified tool interface, translating tool calls into runtime operations and managing VLA backends, runtimes, and acceleration presets within a stable inference boundary. The data plane handles continuous observation and command streams using standard robot middleware, specifically ROS2 in our implementation. In this design, the control plane determines what to start, stop, invoke, or switch, while the data plane carries the observation and command streams consumed by those runtime components. This separation ensures that task logic remains independent of transport protocols or backend-specific code, allowing the same interface to support both low-latency direct commands and persistent closed-loop execution. Figure~\ref{fig:framework} summarizes this internal architecture, including tool invocation from the agent side, unified orchestration, capability exposure, the direct and PIC execution paths, and backend-preset injection at the inference module.

\subsection{Overview}
\label{sec:method-overview}

RoboNeuron structures robot capabilities as agent-callable tools and facilitates two distinct execution paths. The direct path processes tool calls containing structured arguments, which the system validates and encodes before publishing a corresponding ROS message to the data plane. This path is intended for one-shot, low-latency primitives such as base velocity updates, discrete triggers, or incremental end-effector movements. In contrast, the closed-loop path initializes long-running modules that communicate via ROS2 topics, establishing a persistent loop that generates actions from streaming observations. Although these paths differ in runtime behavior, they are exposed through the same agent-facing interface. This allows the agent to select an execution mode based on task needs without switching to a different control surface or learning a separate invocation pattern.

All orchestration is managed at runtime by the agent through tool calls. RoboNeuron does not embed a task planner of its own; instead, it is designed to be orchestrated by an external LLM agent through the same tool interface. In this way, it provides the runtime infrastructure to start, stop, and manage these processes, ensuring that closed-loop execution is handled as an explicitly governed system service rather than an unmanaged background task. This design keeps high-level decision making outside the middleware while giving the agent direct control over execution-side lifecycle operations.

Process isolation is a critical requirement for stable deployment. Each long-running module operates as a standalone ROS2 node within a separate OS process, initiated via the \texttt{spawn} method. This design avoids common issues related to fork-safety in \texttt{rclpy} environments and ensures that process termination is explicit. A dedicated stop tool manages process cleanup, utilizing a bounded wait period before resorting to a forced termination to maintain predictable system states. As a result, long-running capabilities remain observable and controllable at runtime rather than being hidden behind ad hoc process management logic.

\subsection{Tool Derivation}
\label{sec:method-tool-derivation}

Interface drift in agentic systems often arises from manual wrappers where tool signatures and robot messages evolve asynchronously, resulting in redundant integration code. In practice, this means that adding or modifying robot capabilities often requires repeated wrapper updates, even when the underlying change is small. Over time, such hand-written interfaces become difficult to maintain consistently across capabilities. To mitigate this, RoboNeuron employs schema-based derivation to generate tool signatures directly from ROS message and interface definitions. This approach ensures that agent-facing schemas remain synchronized with robot I/O as APIs change, while allowing newly added robot capabilities to be exposed through the same derivation path instead of requiring separate wrappers for each interface.

The system parses ROS message definitions to resolve fields, including nested structures and arrays, and constructs a structured tool argument schema. This schema is implemented as a typed model with built-in validation. Tool calls are converted into ROS messages by recursively mapping validated arguments to the message fields. Each derived tool is bound to a publisher on a designated topic, ensuring that calls result in deterministic message delivery. Furthermore, the interface supports the publication of short command sequences with specified step durations, enabling simple scripted motions while maintaining a consistent schema. This makes message-level capabilities easier to reuse across tasks, since the agent always interacts with a stable tool surface even when the underlying robot message types differ. Our current implementation focuses on message-level exposure via topics, which addresses the low-latency primitives required for our evaluation. Algorithm~\ref{alg:schema_tool} summarizes this schema-based derivation pipeline, from ROS message parsing and typed schema construction to validated invocation and ROS message publication.

\begin{algorithm}[t]
\caption{Schema-based Tool Derivation and ROS Message Invocation}
\label{alg:schema_tool}
\begin{algorithmic}[1]
\Require Message schema $S$, publish endpoint $e$ (topic/type), tool name $n$
\Ensure Exposed tool $T$
\State Parse $S$ and resolve fields (including nested/array fields)
\State Build tool argument schema $\Sigma$ from resolved fields
\State Build encoder $E:\Sigma \rightarrow \texttt{message}$ and bind publisher $P$ to endpoint $e$
\State Register tool $T \triangleq (n,\Sigma,E,P)$ to the tool registry
\Statex
\State On each call: validate $\textit{args}$ against $\Sigma$, encode $m \gets E(\hat{\textit{args}})$, dispatch $m$ via $P$
\end{algorithmic}
\end{algorithm}

\subsection{PIC Modules}
\label{sec:method-pic}

Closed-loop execution is facilitated through a perception--inference--control (PIC) composition. This framework establishes clear responsibilities for each of the three modules, connecting them through topic-based streams on the data plane to form a reusable execution contract. The contract is simple: visual observations flow from perception to inference, action vectors flow from inference to control, and executable robot commands are emitted by the controller. By keeping this contract fixed, the system can vary the internal implementation of the inference module without requiring changes to the surrounding perception or control interfaces.

The perception module manages the conversion of environment data into a subscribable visual stream. It utilizes either a camera driver or a simulator interface as the source, continuously publishing \texttt{sensor\_msgs/Image} messages to the data plane. This module is restricted to data acquisition and formatting, avoiding any reasoning or action generation logic. The inference module operates within the stable inference boundary, subscribing to the image topic and converting frames into the required model format. Given the current instruction, it queries the selected VLA backend and produces an action under a fixed semantic contract: a 6-DoF end-effector delta in position and orientation together with a gripper command. For transport on the ROS2 data plane, this action is flattened into a vector and published as \texttt{std\_msgs/Float64MultiArray}. This keeps the transport format simple while preserving a stable action meaning across different inference implementations.

The control module subscribes to this topic and interprets the vector under the same action contract rather than as an arbitrary numeric array. Upon initialization, the controller parses a URDF (omitting visual and collision elements for efficiency) to build a kinematic chain, and it subscribes to both the action topic and joint-state feedback. Using the current joint state, it first computes the current end-effector pose through forward kinematics, applies the predicted delta action to obtain the target pose, and then solves the inverse kinematics (IK) with the current joint state as the initial guess. Commands are finally issued as joint-space trajectories or states, depending on the runtime configuration. Because the controller only depends on the fixed action contract and the ROS topic bindings, the same control-side logic can be reused even when the VLA backend or inference runtime changes inside the stable boundary.

\subsection{Inference Switching}
\label{sec:method-switching}

RoboNeuron localizes VLA-specific logic within the inference boundary to enable topology-preserving switching. The inference module selects a VLA backend through a model wrapper registry, where checkpoints can be specified at startup or resolved via project configurations. Runtime and acceleration settings are injected through the same boundary during initialization. In the current prototype, this mechanism covers OpenVLA\cite{OpenVLA}, OpenVLA-OFT\cite{OpenVLA-OFT}, and $\pi_0$\cite{pi0} backends, SGLang\cite{SGLang} as an alternative serving runtime, and FastV-based acceleration presets~\cite{fastv}. These options differ in internal implementation, serving logic, and optimization behavior, but they are integrated under the same runtime abstraction.

These internal configurations do not alter the external I/O contract, which maintains a fixed image input and a vector-based action output. Consequently, switching VLA backends or acceleration settings does not require modifications to the perception or control modules, nor does it disrupt the underlying topic connections. The rest of the system continues to see the same image topic, the same action topic, and the same downstream controller behavior. This modularity reduces redundant integration work and ensures that cross-model comparisons are not confounded by changes in the system wiring.

\subsection{Workflows}
\label{sec:method-workflows}

The system supports two primary workflows. In the direct workflow, the agent invokes a message-publishing tool with validated arguments to trigger immediate one-shot effects. This path bypasses the VLA inference module and is suitable for high-frequency primitives or timed command sequences. In the closed-loop workflow, the agent initializes the PIC modules, establishing a continuous data stream from visual perception to action generation and joint control. The perception module provides the visual stream, the inference module generates actions within the stable boundary, and the control module executes these actions via URDF-based kinematics. The two workflows therefore share a common orchestration surface while differing in whether execution is immediate and message-driven or persistent and stream-driven. If a change in the VLA backend or acceleration preset is required, the agent can restart the inference module with the new configuration while the perception and control modules continue to run, ensuring the overall system topology remains intact. Algorithm~\ref{alg:workflow_switch} presents this closed-loop orchestration procedure and the corresponding topology-preserving inference-switching routine.

\begin{algorithm}[t]
\caption{Closed-loop PIC Orchestration with Topology-preserving Inference Switching}
\label{alg:workflow_switch}
\begin{algorithmic}[1]
\Require Instruction $\tau$, topics $(t_{\mathrm{img}}, t_{\mathrm{act}}, t_{\mathrm{cmd}})$, backend choice $b$, accel preset $p$
\State Start perception on $t_{\mathrm{img}}$ and start control consuming $(t_{\mathrm{act}})$ and publishing $t_{\mathrm{cmd}}$
\State Start inference with $(b,p)$, subscribing to $t_{\mathrm{img}}$ and publishing actions to $t_{\mathrm{act}}$
\State Agent monitors progress and may request termination through stop tools
\If{switch requested from $(b,p)$ to $(b',p')$}
    \State Stop inference
    \State Start inference with $(b',p')$ using the same $(t_{\mathrm{img}}, t_{\mathrm{act}})$
\EndIf
\State Stop inference, stop control, stop perception
\end{algorithmic}
\end{algorithm}

\section{EXPERIMENTS}
\label{sec:experiments}

\subsection{Experimental Setup}
\label{sec:exp-setup}

We evaluate RoboNeuron as a middle-layer infrastructure for agentic VLA deployment. Accordingly, our experiments focus on the engineering roles identified in the introduction: scalable interface bridging, modular organization of long-running behaviors, and topology-preserving switching across backend-side implementations. The evaluation is organized into two parts. Cases~I--III are mechanism-oriented case studies that validate the unified tool interface, the two execution paths, and explicit lifecycle control across both simulation and real hardware. Case~IV is a controlled benchmark evaluation that studies backend/runtime variation inside the stable inference boundary while keeping the surrounding execution topology fixed.

All experiments follow the two execution modes described in Sec.~\ref{sec:method}. In the direct path, a tool call is validated, encoded, and published as a deterministic ROS~2 message for one-shot execution. In the closed-loop path, perception--inference--control (PIC) modules run as persistent ROS~2 nodes and are explicitly started, stopped, and restarted through agent-side tool calls. This setup allows us to examine both immediate message-driven behaviors and longer-running stream-driven behaviors under the same orchestration surface, rather than treating them as separate control systems.

Our experiments cover both simulation and physical hardware. In simulation, we study multi-platform base control and a single-arm execution stack to validate direct tool invocation and execution-side control adaptation. On real hardware, we deploy RoboNeuron on an FR3 arm for a vision-guided grasping task and focus on runtime semantics, especially explicit lifecycle control for persistent closed-loop execution. For backend switching and acceleration analysis, we use the LIBERO benchmark~\cite{LIBERO}, specifically LIBERO-Spatial, LIBERO-Object, LIBERO-Goal, and LIBERO-Long. Each suite contains 10 tasks, and each task is evaluated over 50 episodes.

Case~IV consists of two complementary benchmark studies. The first examines OpenVLA-OFT~\cite{OpenVLA-OFT} under pruning variants, where only the pruning preset inside the inference module is changed while the rest of the evaluation harness is kept fixed. The second examines OpenVLA~\cite{OpenVLA} under runtime and pruning variants through single-step inference latency, isolating inference-side efficiency changes when additional runtime variation is introduced. This design separates task-level comparison under a fixed serving setup from inference-level efficiency comparison under runtime-augmented configurations, while keeping the external observation/action contract unchanged.

\subsection{Case Studies (Case I--III)}
\label{sec:exp-cases}

The following case studies examine RoboNeuron's core mechanisms rather than task-level leaderboard performance. They focus on whether robot capabilities can be exposed under a unified interface, whether different execution patterns can be orchestrated through the same runtime layer, and whether long-running closed-loop behaviors can be managed explicitly and safely.

\subsubsection{Case I: Multi-platform Direct Tool Invocation}
\label{sec:exp-case1}

\begin{figure}[htbp]
\centering
\includegraphics[width=\columnwidth]{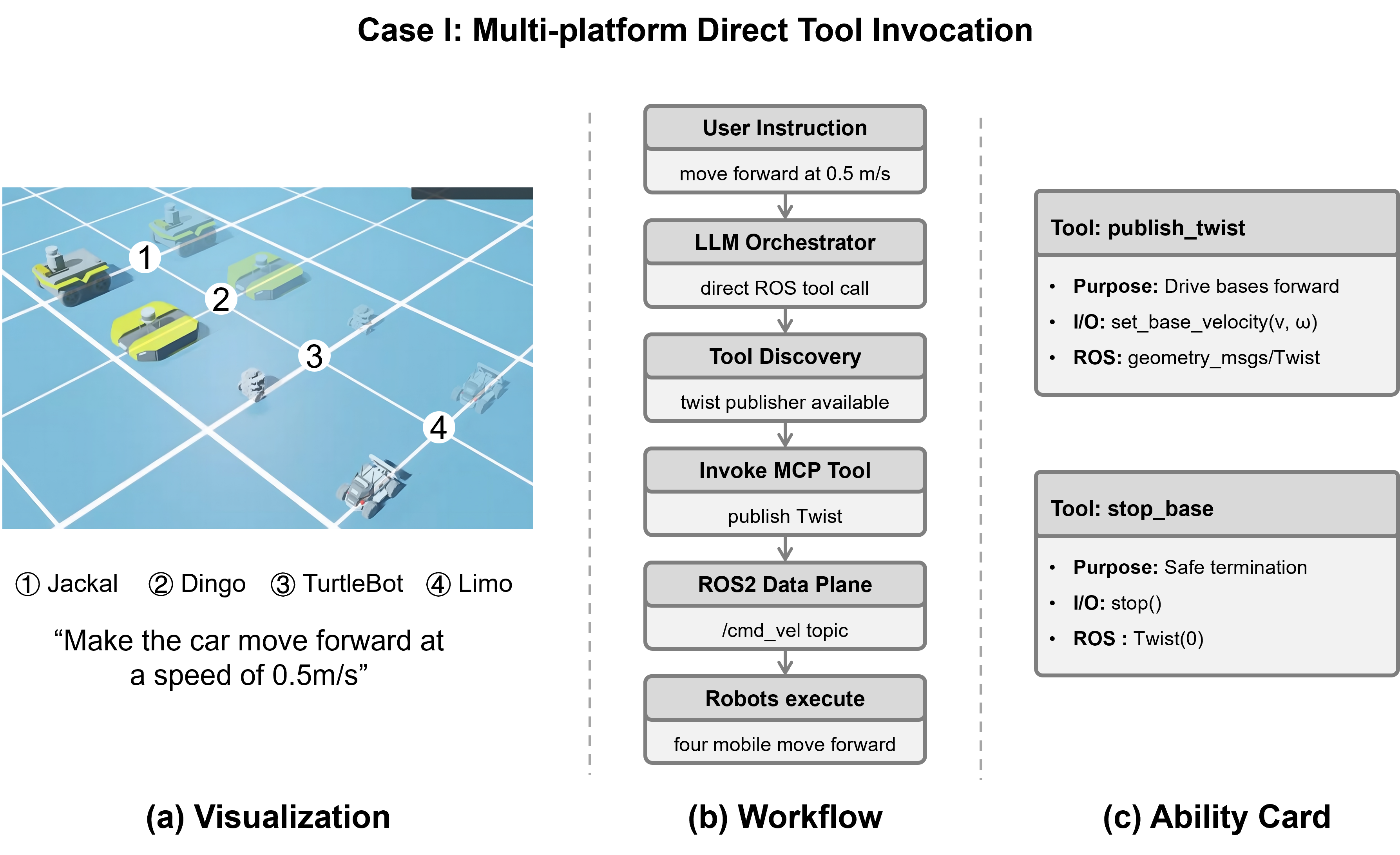}
\caption{Case~I: multi-platform direct tool invocation in simulation. A shared velocity-command tool is reused across multiple mobile platforms through the direct path.}
\label{fig:case1}
\end{figure}

We first examine the direct path in a simulation scenario where a single agent commands multiple mobile platforms. This case highlights interface reuse under a unified tool surface: the same structured velocity-command tool can be bound to different platform-specific controllers without requiring separate wrappers for each robot. RoboNeuron converts each tool call into a command message on the data plane, providing a direct and low-overhead control path. Figure~\ref{fig:case1} illustrates this workflow, from tool discovery and MCP invocation to ROS2 command publication and multi-platform execution.

\subsubsection{Case II: Single-arm Execution Stack via Unified Capability Interface}
\label{sec:exp-case2}

\begin{figure}[htbp]
\centering
\includegraphics[width=\columnwidth]{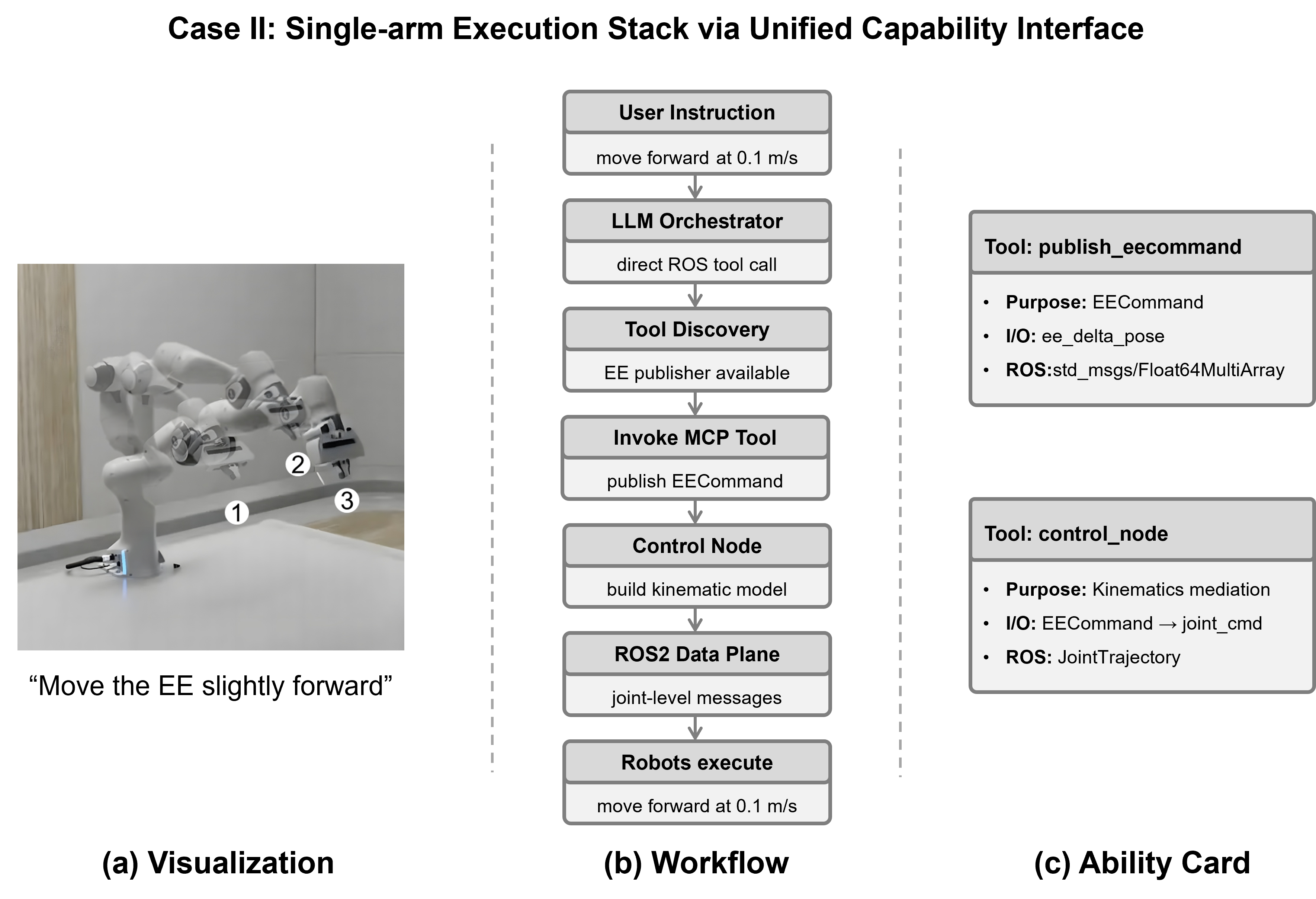}
\caption{Case~II: single-arm execution stack under a unified capability interface. Structured arm-motion tools are executed through ROS2 with control-side adaptation.}
\label{fig:case2}
\end{figure}

We next validate the execution stack in a simulated single-arm setup, showing that the same agent-facing interface can cover both message-level capability exposure and control-side adaptation. In the direct path, the agent invokes structured tools that map to robot I/O, such as joint targets or trajectories. To support embodiment-specific constraints, the control module constructs a kinematic chain from the URDF at runtime and converts high-level end-effector commands into executable joint-space commands dispatched via ROS2. This design keeps the invocation interface stable while allowing execution-side logic to handle robot-specific control details. Figure~\ref{fig:case2} shows this single-arm execution stack, including direct tool invocation, control-side mediation, and ROS2 joint-level command execution.

\subsubsection{Case III: Closed-loop VLA Manipulation on Real Hardware}
\label{sec:exp-case3}

\begin{figure}[htbp]
\centering
\includegraphics[width=\columnwidth]{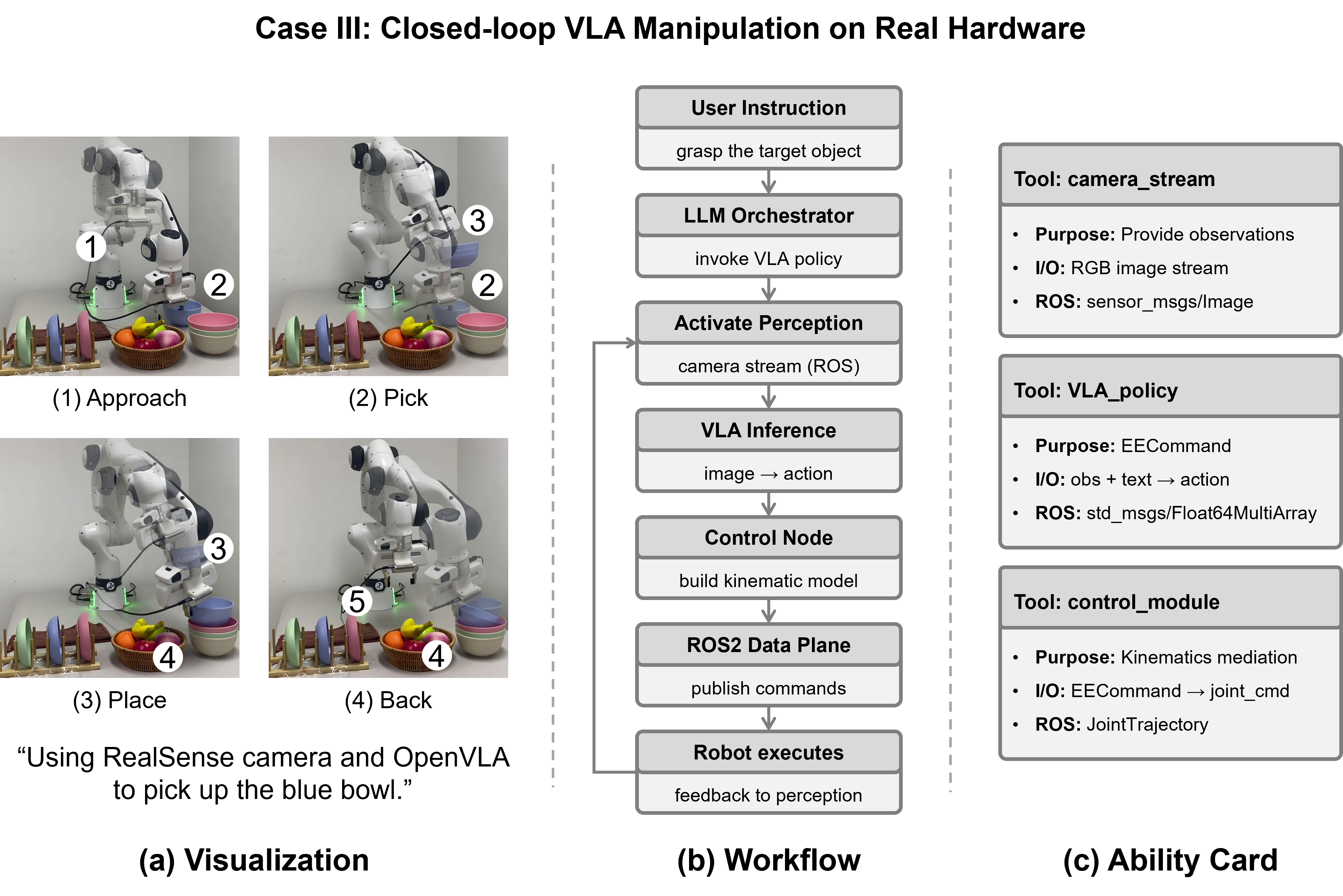}
\caption{Case~III: closed-loop VLA manipulation on real hardware. An FR3 grasping task runs through perception, VLA inference, and control under explicit lifecycle management.}
\label{fig:case3}
\end{figure}

To verify closed-loop execution on real hardware, we deploy RoboNeuron on an FR3 arm for a vision-guided grasping task. The agent initializes a PIC capability in which the perception module streams images, the inference module produces actions inside the stable boundary, and the control module dispatches executable commands to the physical robot. The emphasis of this case is not large-scale task statistics but runtime semantics: the closed-loop behavior is started, supervised, and terminated through explicit lifecycle control rather than being left as an unmanaged background process. Figure~\ref{fig:case3} presents both the closed-loop execution chain and a representative FR3 grasp sequence under explicit lifecycle management.

\subsection{Case IV: Topology-preserving Switching and Acceleration Integration}
\label{sec:exp-case4}

Case~IV evaluates whether RoboNeuron can vary backend-side implementations without changing the surrounding execution topology. Throughout this study, the observation stream, action transport format, downstream control adapter, and ROS2 topic bindings are kept fixed. All variation is confined to the inference module inside the stable boundary, so differences in performance can be attributed to backend-side changes rather than preprocessing, controller wiring, or external system restructuring.

\paragraph{\textbf{OpenVLA-OFT pruning variants}}
Our first study measures task success under pruning-only variants built on top of OpenVLA-OFT~\cite{OpenVLA-OFT}. Here, the serving setup is fixed and only the FastV-based pruning preset~\cite{fastv} inside the inference module is varied. Table~\ref{tab:libero_pruning} reports success rates on the four LIBERO suites together with normalized pruning-only inference speedup relative to the OpenVLA-OFT baseline. This experiment is intended to test whether moderate model-side changes can be compared under a fixed harness without altering the surrounding perception or control pipeline.

\begin{figure}[t]
\centering
\includegraphics[width=\columnwidth]{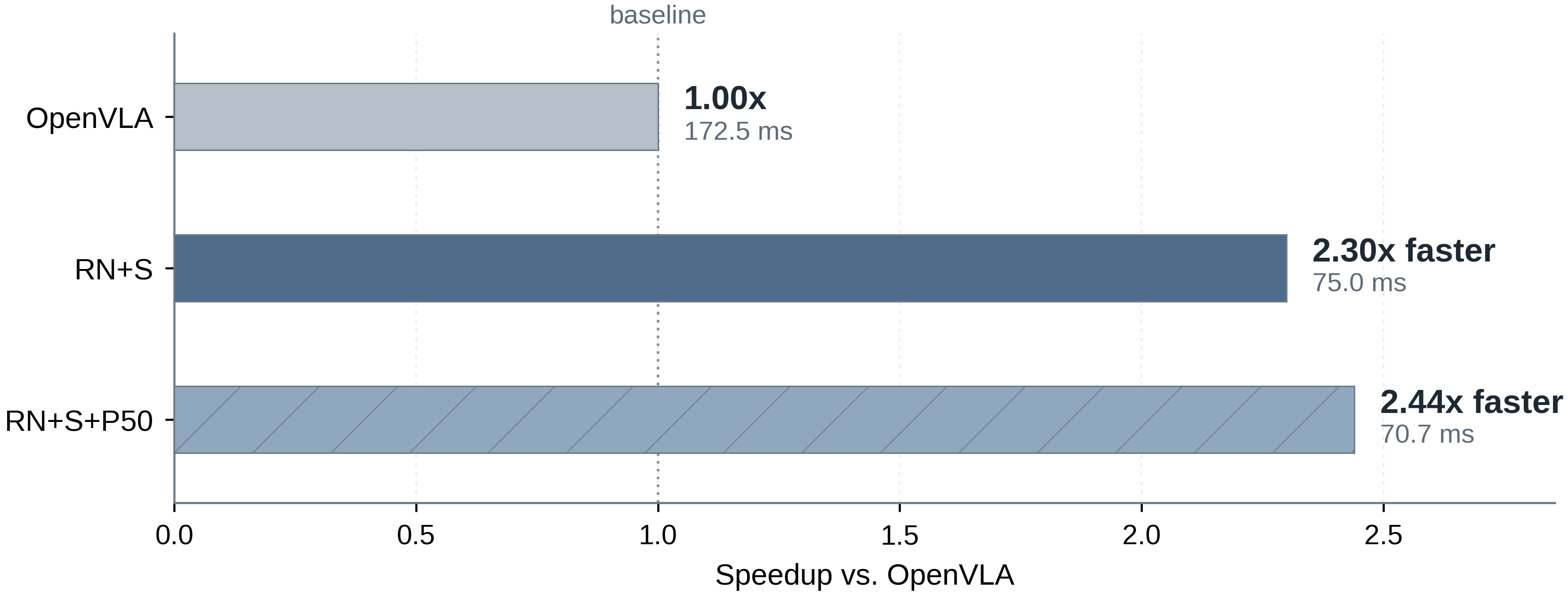}
\caption{Single-step VLA inference latency on RTX~4090 (ms; lower is better) and normalized inference speedup relative to OpenVLA. RN denotes RoboNeuron overhead, S denotes the SGLang serving runtime, and P50 denotes 50\% pruning.}
\label{fig:case4_speedup}
\end{figure}

\paragraph{\textbf{OpenVLA runtime and pruning variants}}
Our second study measures single-step inference latency on top of OpenVLA~\cite{OpenVLA}. Unlike Table~\ref{tab:libero_pruning}, this setup introduces runtime variation in addition to pruning. In particular, we compare the baseline OpenVLA path, RoboNeuron integration overhead, SGLang-based serving~\cite{SGLang}, and runtime-augmented pruning variants. Figure~\ref{fig:case4_speedup} reports latency and normalized speedup on an RTX~4090. This experiment is used to isolate inference-efficiency trends under runtime-aware configurations rather than task-level success under a fixed serving stack.

Taken together, these two studies show that RoboNeuron supports controlled comparison of different inference implementations while preserving the same external observation/action contract and the same surrounding system topology.

\begin{table*}[t]
\centering
\caption{LIBERO success rate (\%) and normalized pruning-only inference speedup relative to OpenVLA-OFT. Results are reported on LIBERO-Spatial, LIBERO-Object, LIBERO-Goal, and LIBERO-Long. P25/P50/P75 denote FastV-based pruning ratios.}
\label{tab:libero_pruning}
\small
\setlength{\tabcolsep}{5.0pt}
\renewcommand{\arraystretch}{1.06}
\begin{tabular}{@{}lccccc@{}}
\toprule
\multirow{2}{*}{Method} & \multicolumn{4}{c}{Succ. (\%, $\Delta$)} & \multirow{2}{*}{Spd. ($\times$)} \\
\cmidrule(lr){2-5}
 & Spa. & Obj. & Goal & Long & \\
\midrule
OpenVLA-OFT
& 98.6 (\same) & 97.6 (\same) & 97.2 (\same) & \textbf{95.8} (\same) & 1.00 \\
RoboNeuron + P25
& 98.8 (\up{0.2}) & 98.0 (\up{0.4}) & \textbf{97.6} (\up{0.4}) & 94.4 (\down{1.4}) & 1.03 \\
RoboNeuron + P50
& \textbf{99.2} (\up{0.6}) & \textbf{98.4} (\up{0.8}) & 96.8 (\down{0.4}) & 94.4 (\down{1.4}) & 1.17 \\
RoboNeuron + P75
& 98.4 (\down{0.2}) & 89.8 (\down{7.8}) & 96.6 (\down{0.6}) & 89.2 (\down{6.6}) & \textbf{1.58} \\
\bottomrule
\end{tabular}
\end{table*}

\subsection{Summary and Limitations}
\label{sec:exp-summary}

Across these experiments, RoboNeuron supports both direct tool invocation and PIC closed-loop execution under a unified interface. The simulation cases show that robot capabilities can be exposed and reused across different execution settings, while the real-hardware case demonstrates that persistent closed-loop behaviors can be managed through explicit lifecycle control. The benchmark study further shows that backend-side variants can be compared inside a stable inference boundary without changing the surrounding topology.

At the same time, the evaluation remains focused on mechanism validation. The real-robot study emphasizes feasibility and runtime semantics rather than large-scale statistical performance. Our current prototype also focuses on topic-based capability exposure and a vector action transport format. Extending the same ideas to broader ROS interface types and more diverse hardware settings remains future work.

\section{CONCLUSIONS}
\label{sec:conclusion}

This paper introduced RoboNeuron, a middleware infrastructure that bridges agent tool calling and robot middleware for VLA deployment. By separating semantics from transport through a dual-plane architecture, RoboNeuron provides a unified MCP tool interface with structured tools derived directly from robot schemas, reducing interface drift between agent-facing signatures and robot I/O. The system offers two execution paths: a direct path for low-latency primitives and a closed-loop path for long-running behaviors with explicit lifecycle management. RoboNeuron also localizes VLA variability within a stable inference boundary, allowing backends and acceleration presets to be switched without altering the surrounding perception and control topology. Simulation and hardware results show that these mechanisms address practical challenges in scalability, system organization, and backend evolution for agentic robotics. Future work will expand interface coverage beyond topic-based exposure and broaden evaluation of latency and hardware performance across more deployment settings.

\addtolength{\textheight}{-12cm}   









\end{document}